# Hand Gestures Recognition in Videos Taken with Lensless Camera

YINGER ZHANG[1], ZHOUYI WU[1], PEIYING LIN[1], YANG PAN[1], YUTING WU[1], LIUFANG ZHANG[2] AND JIANGTAO HUANGFU[1,*]

[1] *Laboratory of Applied Research on Electromagnetics, Zhejiang University, Hangzhou, 310027, China*
[2] *Kunming Cotech Communication & Information Systems Co., Ltd, Kunming 650106, China*
*\*huangfujt@zju.edu.cn*

**Abstract:** A lensless camera is an imaging system that uses a mask in place of a lens, making it thinner, lighter, and less expensive than a lensed camera. However, additional complex computation and time are required for image reconstruction. This work proposes a deep learning model named Raw3dNet that recognizes hand gestures directly on raw videos captured by a lensless camera without the need for image restoration. In addition to conserving computational resources, the reconstruction-free method provides privacy protection. Raw3dNet is a novel end-to-end deep neural network model for the recognition of hand gestures in lensless imaging systems. It is created specifically for raw video captured by a lensless camera and has the ability to properly extract and combine temporal and spatial features. The network is composed of two stages: 1. spatial feature extractor (SFE), which enhances the spatial features of each frame prior to temporal convolution; 2. 3D-ResNet, which implements spatial and temporal convolution of video streams. The proposed model achieves 98.59% accuracy on the Cambridge Hand Gesture dataset in the lensless optical experiment, which is comparable to the lensed-camera result. Additionally, the feasibility of physical object recognition is assessed. Furtherly, we show that the recognition can be achieved with respectable accuracy using only a tiny portion of the original raw data, indicating the potential for reducing data traffic in cloud computing scenarios.

## 1. Introduction

Online-connected camera systems are becoming more widespread in modern society. These systems use cameras to take pictures, and then send them to the cloud for task-specific computational analysis. These systems are vulnerable to hacking, resulting in data leakage and privacy loss. Lensless cameras [1,2,3,4,5] can offer optical-level privacy protection because they encode the scene into a pattern that is not human-interpretable. This means that even if the captured data is leaked during transmission, the attacker cannot decipher its true meaning. However, there are some drawbacks to lensless cameras. In addition to using more computational resources, image reconstruction also introduces errors and artifacts, and the quality of the restored images is invariably worse than the lensed one. Fortunately, many scenes today concentrate on a specific visual task, such as object location, face recognition, text recognition, etc., making it unnecessary to acquire scene-resembling imaging. Studies have suggested performing visual tasks directly on the raw data from lensless cameras to reduce computational burdens. [6,7] directly applied object recognition on the encoded pattern. [8] proposed a lens-free coded aperture camera system for human action recognition without image restoration.

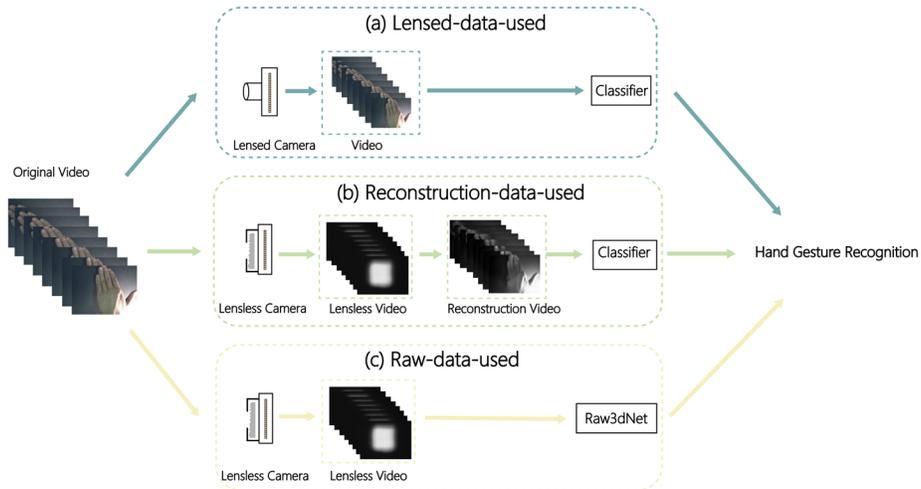

Fig. 1. Comparison of the methods for recognizing hand gestures that use raw data, lensed data, and reconstruction data.

In this research, we propose Raw3dNet, a deep learning model to directly recognize hand gesture on raw videos taken by a lensless camera. Fig. 1 compares the proposed system (raw data used) with lensed camera (lensed-data-used), and lensless camera (reconstruction-data-used) in hand gesture recognition. To our knowledge, Raw3dNet is the first end-to-end deep neural network model for hand gesture recognition in a lensless imaging system. In order to properly extract and combine the temporal and spatial features from raw video taken with a lensless camera, Raw3dNet is created specifically for this purpose. Raw3dNet is divided into two phases: 1. Prior to performing temporal convolution, the spatial features of each frame are intensified by the spatial feature extractor (SFE). 2. 3D-ResNet implements spatial and temporal convolution of video streams. Lensless optic experiments demonstrate that there is enough information in the raw data to recognize hand gestures. In terms of predictive accuracy, the proposed raw-data-used method performs better than the reconstruction-data-used method and is comparable to the lensed-data-used method. Additionally, we show that the recognition can be achieved using only a small portion of the original raw data, indicating the potential for reducing data traffic in cloud computing scenarios. The proposed system is anticipated to be used for inference tasks that are sensitive to privacy in cloud computing scenarios due to its optical encryption, low computational resource consumption, and small data requirement.

## 2. Related work

### 2.1 Lensless imaging

With the development of computational imaging, several small-format lensless cameras [1,2,3,4,5,9,10] were introduced, and they had a wide range of applications [6,7,11,12,13]. In this study, DiffuserCam [5,10] served as the optical hardware system, where the lens was replaced with a phase mask positioned in front of the bare sensor at a close proximity. To recover the image from the encoded pattern, a lensless camera required additional reconstruction techniques. And there were three main categories of image reconstruction algorithms: the iterative optimization method, which employed convex optimization like gradient descent (GD) and alternating direction methods of multipliers (ADMM) iteratively [14]; the pure end-to-end convolutional neural network [15,16,17,18]; and the unrolled optimizer, which addressed the inverse problem by incorporating the system model into the network [19]. Since the reconstruction is not essential for inference tasks, it would be advantageous if it could be avoided.

### 2.2 Reconstruction-free visual inference

In a variety of fields, reconstruction-free recognition has been used. Lens-based optical hardwares, as opposed to the Diffusercam used in this work, were used in [20,21,22]. The frameworks suggested in [23,24,25] did not include lenses but were intended for laser-illuminated targets. To lower measurement rates, reconstruction-free recognition has been added to single-pixel cameras [26,27]. In a mask-based lensless camera that was similar to DiffuserCam [6,7,8], reconstruction-free recognition was also implemented. For privacy-preserving action recognition in particular, [8] used lensless optical hardware, and before classification, it must compute non-invertible motion features between pairs of frames. In contrast to this approach, Raw3dNet in the work is a more user-friendly end-to-end deep neural network model.

*2.3 Hand gesture recognition*

Vision-based hand gesture recognition techniques include both static and dynamic gesture-oriented approaches. In this paper, we discuss the dynamic hand gesture recognition, which has a temporal aspect. Many methods based on deep learning have been proposed in the field of action or gesture recognition. Methods based on two-stream convolutional networks [28], methods based on three-dimensional convolutional networks [29], and methods based on long short-term memory (LSTM) [30] are the principal representative works.

## 3. Method

Fig. 1 (c) depicts the framework. Lensless video is captured by a lensless camera and encoded into raw data. In a subsequent step, all frames in the lensless video are simultaneously fed into an end-to-end neural network that recognizes hand gestures without any image reconstruction. In Section 3.1, the lensless camera system in this study are introduced. In Section 3.2, the datasets used in the study are described. In Section 3.3, we explain the Raw3dNet consisting of SFE and 3D-ResNet.

*3.1 Lensless imaging system*

In this study, a lensless camera system is built by using the hardware prototype of DiffuserCam as the lensless camera system. In accordance with Fig. 2, the lensless camera is made up of a diffuser (luminit 1°) with a 1/4-sensor-size aperture that is positioned 2mm away from a sensor (ov5647, 5-megapixel), and a 3D-printed bracket is used to maintain the diffuser's position. A 1080p LED monitor is used to display the target videos, which are physically 16cm×20cm, at a distance of 25 cm from the camera to match the lensless camera's field of view. The monitor displays L frames of a video in order, and the sensor then separately recorded L frames of raw data to create a lensless video. In the physical object experiment, the tester's hand is placed at the position of LED monitor.

By presuming that all points in the scene are incoherent and shift invariant, the following Eq. (1) serves as the lensless camera system forward model [5]:

$$\mathbf{b}(x,y) = crop[\mathbf{h}(x,y) * \mathbf{x}(x,y)] = \mathbf{CH}\mathbf{x} \tag{1}$$

Here **b** denotes the sensor measurement; **h** represents the system PSF (the point spread function), which can be estimated through calibration; **x** represents target scene; $(x,y)$ is the coordinates; * denotes 2D convolution. Given the finite size of the sensor, a crop operation is applied to the model. ***C*** and ***H*** express crop operation and PSF, respectively, under matrix-vector notation.

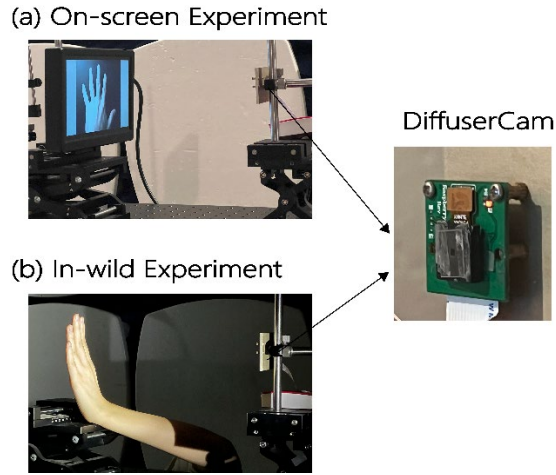

Fig 2. Diagrammatic overview of two experiments. (a) A monitor is placed in front of the lensless camera and shows the videos. (b) Comparative experiment of placing hands in front of the lensless camera.

### 3.2 Dataset

Two kinds experiments and datasets are implemented in the work, Fig. 2 (a) shows the data collection in on-screen experiments, while Fig. 2 (b) shows the data collection in in-wild experiments. The on-screen dataset is a public available dataset named Cambridge Hand Gesture [31], on which we trained and test Raw3dNet. The in-wild dataset is constructed by the lensless videos we took of actual, physical hands with three-dimensional structures. We just test on the in-wild dataset to check the robustness and generalizability of the model trained using the on-screen dataset.

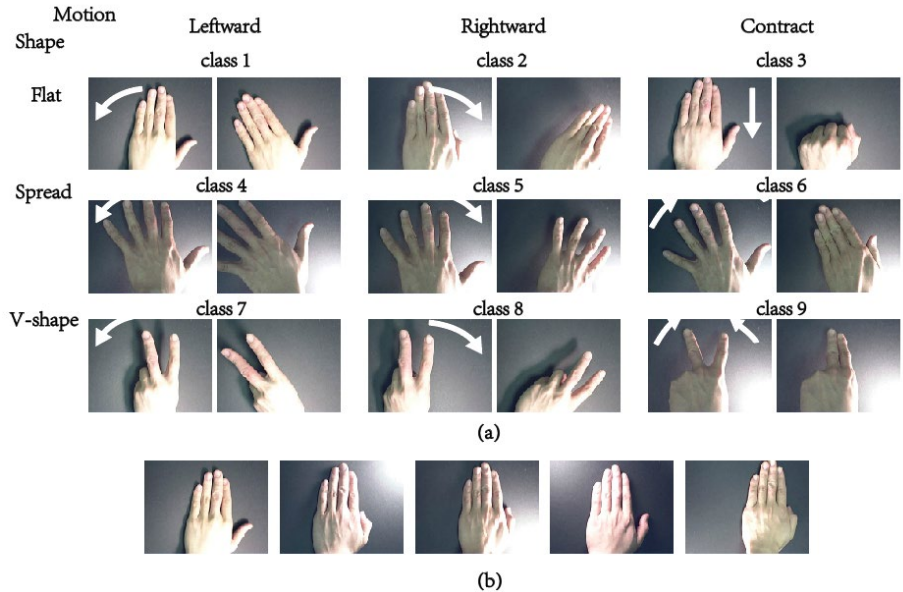

Fig. 3. Description of on-screen dataset. (a) The dataset contains 9 gesture classes. (b) 5 different illumination prototypes for better simulation of the real world.

As seen in Fig. 3 (a), Cambridge Hand Gesture is made up of 900 image sequences representing 9 different gesture classes, each of which is indicated by 3 primitive hand shapes and 3 primitive motions. As a result, the goal task for this dataset is to categorize various shapes and motions simultaneously. Additionally, there are five different illumination prototypes, as seen in Fig. 3 (b), for a more accurate representation of the real world, which creates a bigger challenge for the lensless imaging system. 20% of the dataset is used as test dataset, while the remaining 80% serves as the train dataset. Data preprocess is used to expand both train and test data respectively: One video is evenly divided into several sub-videos in form of 8×240×320, where 8 is the length and 240×320 is the size. Every original video sequence contains around 50-70 images. For better use of the data, we divide one video into several sub-videos which consist 8 frames. For example, the first sub-video may contain 1th, 7h, 18th, 25th, 32th, 39th, 46th, 53th images of the original video, and the second sub-video may contain 2th, 8h, 19th, 26th, 33th, 40th, 47th, 54th images of the original video. The sub video number and the step gap between images is somewhat random depending on the image number of the original video. Finally, the train dataset has 2832 videos while the test dataset has 780 videos. We gather the lensless hand gesture video dataset using the lensless imaging device mentioned in section 3.1. The "original video" is the video that is played on the monitor; the "lensless video" is the video that is recorded by the lensless camera.

*3.3 Raw3dNet*

In the fields of action recognition and hand gesture recognition, numerous deep learning-based action representation techniques have been proposed. Methods based on two-stream convolutional networks [28], methods based on 3D convolutional networks [29], and methods based on long short-term memory (LSTM) [30] can be summed up as the main representative works. Various techniques are utilized to capture movement between frames. In a two-stream convolution network, optical flow is created by a set of displacement vector fields between adjacent pairs of frames, t and t+1. It can effectively capture motion between frames in this way. Calculating optical flow for frames taken by a lensless camera, however, is pointless because each sensor pixel collects multiplexed light from widely spread points in the scene. In the end, the 3D convolutional networks approach is selected as the baseline model due to its superior performance to LSTM in the pre-experiments.

Direct extraction of spatial-temporal features from videos is possible with spatial-temporal 3D kernels (3DCNNs) [32]. An enhanced version of the ResNet-based 3DCNN structure is called 3D-ResNet [33]. Shortcut connections that skip a signal from one layer to the next are introduced by ResNet. It is easier to train very deep networks because connections pass through gradient flows of networks from later layers to early layers. The architecture of 3D-ResNet and the residual unit are depicted in Fig.4(d). The size of the input data is 1×8×240×320. Downsampling of the inputs is performed by 3Dconv of size 7×7×7 and three consecutive residual units when the feature maps increase. A predicted vector for 9 categories is the output. We first use the original video and the lensless video as our baseline to train two 3D-ResNet models. As shown in Table 3, the original videos' Top1 classification accuracy of 99.36% represents the upper bound of what we can anticipate. Lower bounds are provided by the performance of 3D-ResNet trained on lensless videos (78.97% accuracy). It is easy to conclude that directly applying 3D-ResNet model to lensless video is ineffective. In order to investigate the cause of the unsatisfactory performance of 3D-ResNet on lensless video, we count the class distribution of predictions and the confuse matrix is displayed in Table 1. We find that the mistakes are more centered on shape prediction than motion prediction. Consider class 1 as an example. Errors tend to be predicted as class 4 or class 7, the images in these classes have different shapes but the same motion as class 1 images. This demonstrates how the model can accurately predict the motion direction but not the object's shape. Both the model and the data contribute to the flaw. From a data perspective, lensless cameras are sensitive to lighting, so uneven lighting in the video degrades the quality of lensless video; From a model perspective, the spatial and temporal convolutions are performed directly on the raw measurements, where each sensor pixel measures multiplexed light from widely dispersed points in the scene, making it difficult to identify the spatial characteristics, ultimately leading to rare enough information being obtained by convolution.

Table 1. Confuse matrix when using 3D-ResNet for lensless video classification.

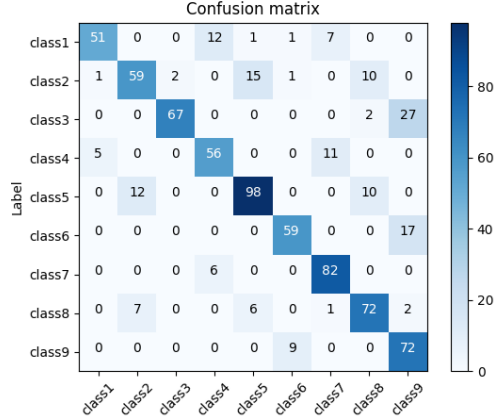

This work suggests adding an SFE (spatial feature extractor) before the 3D-ResNet in order to enhance the spatial features of each frame prior to performing temporal convolution. Encoder-decoder [34] is a frequently used structure for image reconstruction and segmentation because of its superior ability to extract spatial features. Inspired by this, we create a convolution neural network with an encoder-decoder architecture as the SFE. As shown in Fig. 4(b), SFE has an up-sampling architecture and a down-sampling architecture connected by a skip connection. Ablation experiments were done to determine the proper number of layers and channels. The SFE and 3D-ResNet are integrated as Raw3dNet in both training and inference phases. Frames of raw videos are sequentially input into the SFE to obtain feature maps, which are then stacked into 8 frames and fed into 3D-ResNet. Both portions of Raw3dNet (the SFE and 3D-ResNet) are trained simultaneously with one pass of the data in an end-to-end manner. Fig.4 shows an illustration of Raw3dNet and a tabular detailing of the network is shown in Appendix.

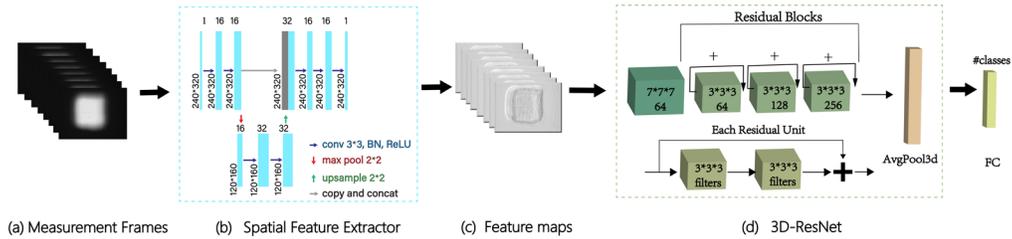

Fig. 4. The architecture of Raw3dNet.

We further validate the significance of SFE with the analysis below. We will examine the improvement of the SFE in shape classification because, as is seen above, predicting errors when using 3D-ResNet for lensless video classification mainly focused on shape classification rather than motion classification. It is widely accepted that the "clustering property" determines the data classification. It reveals the fact that objects belonging to the same category are closer together than those belonging to a different category. An assessment of category clustering effects is based on inter- class correlation and intra-class correlations, and intra-class correlation is larger than inter-class correlation. Typically, the data projected into a high-dimension domain is used to calculate the distance and correlation. A popular and efficient technique for transferring images into high-semantics space is the use of neural networks [35]. In this work, input images are mapped to 512-dimensional vectors using a VGG-16 that has been pretrained on ImageNet, and the correlation between the two input images is determined by the cosine similarity of the 512-dimensional vectors. For evaluation, we select the first frame of each video from class 1, class 4, and class 7, which depicts three different hand shapes (flat, spread and v-shape). Using class 1 as an example, we use the aforementioned technique to determine three correlation values for each image of that class: the average intra-class correlation (with images of class 1) and the average inter-class correlations (with images of class 4 and with images of class 7 respectively). Every class 1 image has a most pertinent category (class 1, class 4, or class 7), and that category has the highest correlation value. The distribution of the class 1's most pertinent category is then counted. The above experiment is conducted on the raw data (Fig. 4. (a)) and feature maps (Fig. 4. (c)) produced by SFE, respectively. Their comparison allows for an evaluation of the SFE's effect. The results in Table 2 suggest that it's difficult to distinguish class 1 from other classes in raw data, resulting in poor shape classification performance. However, after SFE processing, the intra-class correlation of feature maps is higher than the inter-class correlations, laying the groundwork for accurate shape classification.

Table 2. The distribution of the most pertinent category for class 1. Row1 represents images of raw data, and Row2 represents feature maps produced by SFE.

| Dataset | Class 1 | Class 4 | Class 7 |
|---|---|---|---|
| Raw data | 25 | 37 | 10 |
| Feature map | 60 | 6 | 6 |

## 4. Experiment

To show Raw3dNet's abilities in reconstruction-free hand gesture recognition, three experiments are carried out. Training implementation is introduced in Section 4.1. The on-screen dataset is used for the experiments in Section 4.2 and 4.3 and the in-wild dataset for the experiment in Section 4.4.

### 4.1 Training setup

All trainings are implemented on one RTX 2080Ti GPU, with pytorch1.0.0 and python 3.6 under Ubuntu18.04. Raw3dNet is trained for 100 epochs by Adam optimizer with $\beta 1 = 0.9, \beta 2 = 0.99$, a weight decay of 0.001, and a mini-batch size of 12. The initial learning rate is $10^{-3}$ and is then decreased linearly to $10^{-5}$. The network parameters are updated based on Cross Entropy loss. 15% of the train dataset is divided as validation dataset, and the best model is chosen according to the accuracy of validation dataset.

### 4.2 Experiment 1

The definition of datasets is shown in Fig. 5. Original video is the name of the dataset that is displayed on the monitor. Lensless video refers to the dataset made up of the raw data that is captured on the sensor. We create reconstructed datasets by performing image reconstruction on lensless video in order to compare the reconstructed-free method with the reconstructed one. One employed image reconstruction algorithm is the alternating direction method of multipliers

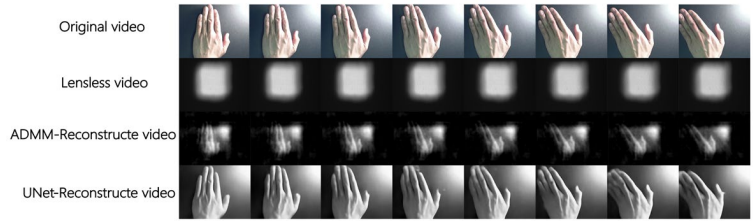

Fig. 5. Definition of datasets

(ADMM), and the associated dataset is named as ADMM-Reconstructed video. UNet-Reconstructed video is another reconstructed dataset in which images are recovered from lensless video using deep neural networks. Five experiments are carried out. We train the 3d-ResNet model on the original video and the reconstructed videos. For comparison, the 3d-ResNet model and Raw3dNet model are trained independently on lensless video. Table 3 lists the dataset, the model, and the predictive accuracy of test dataset. The following three conclusions are drawn from the findings:

First, as shown in Table 3, comparison of Exp4 and Exp5 is used to examine the impact of SFE for lensless video. According to section 3.3, SFE enlarges the spatial feature prior to temporal convolution, boosting hand gesture recognition's predictive accuracy in lensless video from 78.97% to 98.59%. Second, lensless video can achieve accuracy (98.59% in Exp5) comparable to that of a lensed camera (99.36% in Exp1) when trained with Raw3dNet. The third finding is that, regardless of whether the video is restored by ADMM or U-Net, recognizing hand gestures directly on lensless video outperforms recognizing on reconstructed video (Exp2 and Exp3). The conclusion is reasonable because mistakes are introduced during the reconstruction process, which may then prevent successful recognition, particularly in uneven lighting conditions. This sort of error is prevented by reconstruction-free hand gesture recognition.

**Table 3. Comparison of performances for 3D-ResNet/ Raw3dNet for lensless video; comparison for lensless video/reconstruction video/lensed video.**

| Index | Dataset | Model | Accuracy on Test Dataset |
|---|---|---|---|
| Exp1 | Original video | 3d-ResNet | 99.36% |
| Exp2 | ADMM-Reconstructed video | 3d-ResNet | 93.33% |
| Exp3 | UNet-Reconstructed video | 3d-ResNet | 95.64% |
| Exp4 | Lensless video | 3d-ResNet | 78.97% |
| Exp5 | Lensless video | Raw3dNet | 98.59% |

## 4.3 Experiment 2

Each sensor pixel in a lensless camera measures multiplexed light from widely dispersed points in the scene after each point in the scene is mapped to a large pattern on the sensor. As shown in Fig. 6, the raw data generated in this manner is redundant. In this section, we investigate whether hand gesture recognition is possible with a small amount of raw data. Instead of reducing the number of frames, data reduction is done on pixel number of each individual frame. The influences of down sample method and sample ratio on recognition accuracy are investigated. The findings demonstrates that hand gesture recognition can be done with high predictive accuracy using only a small portion of lensless video.

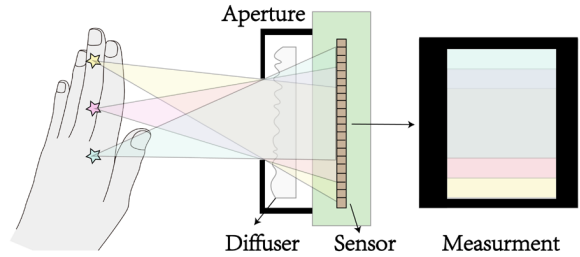

Fig 6. Each point in the scene maps to an encoded pattern on the sensor.

In Table 4, four different down sample techniques are depicted. To enable fair comparison, all down sampled images have 100*75 valid pixels. The down sample method is described as follows: "Resize" means using the bilinear interpolation method to reduce the image from (320,240) to (100,75); "Uniform sample" means cropping the middle part of the raw data

(300,225) and then evenly sampling in steps of 3 to get the (100,75) image; "Random sample" means randomly selecting 7500 pixels from the (320,240) image, not necessarily in a continuous block, and arbitrarily arranging them to form a new (100,75); "Erase" means first resize the raw data to (200,150) and then erase a random subset of the pixels (75%), which means set them to zero, and the remaining 25% of the pixels retain the original position and value information.

As shown in Table 4, "Resize" has the highest accuracy (98.46%), and when compared to raw data of (320,240), the same performance is achieved using only 9.76% of the data. "Uniform sample" also achieves 96.92% accuracy, which is satisfactory. The two down sampling techniques mentioned above only lower the sensor's resolution while maintaining the relative position relationships between pixels in the original data. But in "Erase", the position's randomness increases, causing a slight drop in accuracy (91.54%). Surprisingly, "random sample" achieves 79.74% task accuracy even though the selected pixels are not physically connected to one another and the randomness is high. It implies that the target's information is spread throughout the entire pattern and it does not matter which part of pattern is used. Additionally, we use the "Resize" method to get images that are (50,37) in size, and the accuracy is 90.13%. According to the findings, accuracy declines as data volume decreases.

Table 4. Assessment for various down-sampling techniques and ratios.

| Pixel Size | Compress Method | Accuracy on Test Dataset |
|---|---|---|
| (320,240) | None | 98.59% |
| (100,75) | Resize | 98.46% |
| (100,75) | Uniform sample | 96.92% |
| (100,75) | Random sample | 79.74% |
| (200,150) | Erase (25% reserved) | 91.54% |
| (50,37) | Resize | 90.13% |

In addition, we explore the valid information contained in the sampled data from the perspective of image reconstruction. Fig.7 shows the UNet-Reconstructed videos of different sampling methods. All the compressed-data involves enough effective information to reconstruct the contour of the image to some extent, though not enough details can be obtained. The reconstruction of rough outline and motion ensures the basic classification accuracy (>75%). However, as shown in Fig.8, in some samples of poor light condition, missing data will lead to shape errors in reconstruction (construct Flat to Spread), which is consistent with the Raw3dNet's classification errors (pred class0 to class3). The above results explain why Raw3dNet with sampling data achieves satisfactory performance but never exceeds the original data.

The findings imply that the lensless camera's raw data is indeed highly redundant, and that only a small portion of the original data can successfully recognize hand gestures. This will significantly lessen the burden of data communication in cloud computing scenarios.

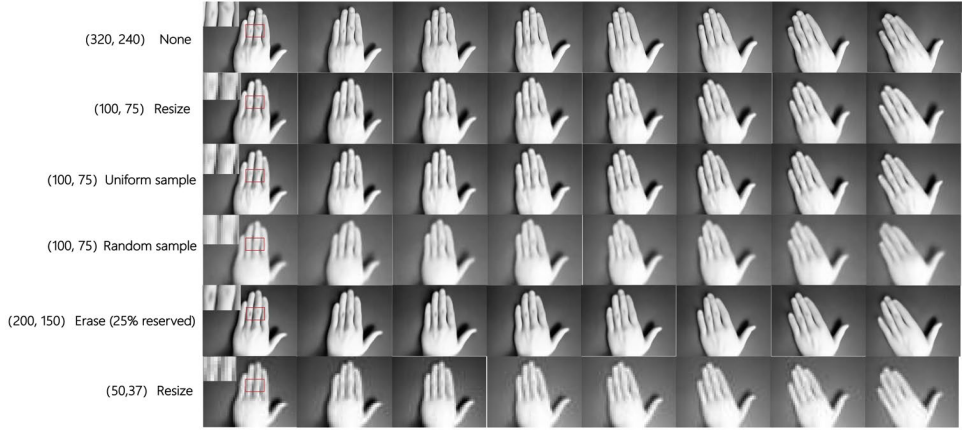

Fig 7. UNet-Reconstructed videos of different down sample method.

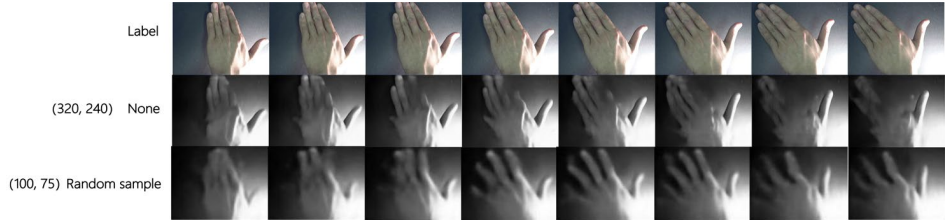

Fig 8. Example of UNet-Reconstructed videos for (320,240) and (100,75) with random sampling.

## 4.4 Experiment 3

We remove the monitor and record lensless videos of actual hand gestures in order to confirm the application effect of Raw3dNet in realistic scenarios. The experiment settings are shown in Fig. 2(b). We use a lensless camera and a real hand that is positioned about the same distance ($\pm 2$cm) as monitor from the lensless camera, a black board is put behind the hands to reduce the disturbance of background, then a white LED is used to light up the scene. The in-wild dataset contains 10 subjects, each performing nine hand gestures under five different lighting conditions, so the in-wild dataset has 450 videos totally. As shown in Fig.9 (a), the LED illuminates 5 different parts of the scene separately to simulate the complex environment: right-top, left-top, left-bottom, right-bottom and center. Fig.9 (b) is a sample of the in-wild dataset.

We test on the in-wild dataset to check the robustness and generalizability of the model trained using the on-screen dataset without any finetuning. The results are shown in Table5. Given that Raw3dNet/3d-ResNet are only trained with on-screen videos and light conditions are different between the two datasets, recognition accuracy decreases in all methods. UNet-reconstructed method has the most dramatic degradation because it utilizes the neural network in both reconstruction and recognition stages where generalization errors accumulate. Lensless video in wild achieves 82.67% accuracy with the trained Raw3dNet, the outcome is still satisfactory and demonstrates that Raw3dNet can generalize beyond monitor-based videos to real-world objects with vastly different lighting conditions. For practical application, assembling a training set of real objects is typically required, but it is always time-consuming. Based on the foregoing conclusion, we can only train the model using the on-screen dataset and extend it to the wild situation, greatly simplifying the training process.

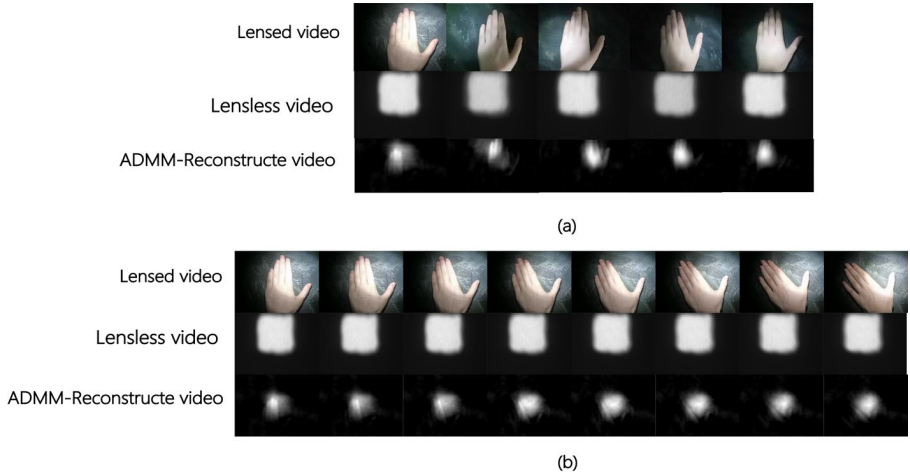

Fig 9. In-wild experiment: the lensed image, lensless iamge and ADMM-reconstructed image. (a). The same gesture under different light condition in the in-wild experiment. (b) a sample video under the same light condition.

Table 5. Performances for lensless video/reconstruction video/lensed video in in-wild dataset.

| Index | Dataset | Model | Accuracy on Test Dataset |
|---|---|---|---|
| Exp1 | lensed video | 3d-ResNet | 386/450 |
| Exp2 | ADMM-Reconstructed video | 3d-ResNet | 352/450 |
| Exp3 | UNet-Reconstructed video | 3d-ResNet | 181/450 |
| Exp4 | Lensless video | Raw3dNet | 372/450 |

## 5. Conclusion

In this work, we propose a reconstruction-free hand gesture recognition method named Raw3dNet for lensless imaging system. In order to provide optical-level protection against data leakage during transmission, the optical system in the work removes the lens and adopts a thin mask to encode the scene into an encoded pattern that is incomprehensible to humans. We concentrate on recognition directly on raw data instead of developing better reconstruction techniques to recover the image from the raw data, which saves computational resources. Raw3dNet is a novel end-to-end neural network model for that task. It is specially designed for raw video recorded by a lensless camera, so that it could properly extract and combine both temporal and spatial features. The optical experiments show that the proposed raw-data-used method performed better in terms of predictive accuracy than the reconstruction-including method and is comparable to the lensed-camera used method. Additionally, we show that the recognition can be achieved using only a small portion of the original raw data, indicating the potential for reducing data traffic in cloud computing scenarios. The proposed system is anticipated to be used for inference tasks that are sensitive to privacy in cloud computing scenarios due to its optical encryption and small data requirement. Whereas, there are some limitations in our work. We just discuss the case with simple background: Images in Cambridge Hand Gesture Dataset have simple and clean background, and to be consistent with training set, the in-wild dataset is obtained in a dark environment to minimize background distraction. Hand gestures recognition in complex and different backgrounds is a common problem in practical applications, and we will study it in the future work.

## Appendix

We outline our Raw3dNet architecture below in Table 6 (Spatial Feature Extractor) and Table 7 (3D-ResNet) for completeness.

Table 6. Architecture of Spatial Feature Extractor

| Layer | Type | Input size | Output size |
|---|---|---|---|

| | Conv3×3, 16, stride 1<br>Batch Normalization<br>Relu | 1×240×320 | 16×240×320 |
| Input layer | | | |
| 2×StackEncoder | Conv3×3, 16, stride 1<br>Batch Normalization<br>Relu | 16×240×320 | 16×240×320 |
| Maxpooling layer | Maxpool 2×2 | 16×240×320 | 16×120×160 |
| 2×StackEncoder | Conv3×3, 32, stride 1<br>Batch Normalization<br>Relu | 16×120×160 | 32×120×160 |
| Upsampling layer | Upsample 2×2<br>Conv3×3, 16, stride 1<br>Batch Normalization<br>Relu | 32×120×160 | 16×240×320 |
| Concat | Concatenate | 16×240×320<br>16×240×320 | 32×240×320 |
| 2×StackDecoder | Conv3×3, 16, stride 1<br>Batch Normalization<br>Relu | 32×240×320 | 16×240×320 |
| Output layer | Conv3×3, 1, stride 1 | 16×240×320 | 1×240×320 |

Table 7. Architecture of 3D-ResNet

| Layer | Type | Input size | Output size |
| --- | --- | --- | --- |
| Conv1 | Conv7×7×7,64, stride (1,2,2)<br>Batch Normalization<br>Relu | 1×8×240×320 | 64×8×120×160 |
| Maxpool | Maxpool 3×3×3, stride 2 | 64×8×120×160 | 64×4×60×80 |
| Layer1 | $\begin{bmatrix} 3\times3\times3, 64, \text{stride } 1 \\ 3\times3\times3, 64 \end{bmatrix}$ | 64×4×60×80 | 64×4×60×80 |
| Layer2 | $\begin{bmatrix} 3\times3\times3, 128, \text{stride } 2 \\ 3\times3\times3, 128 \end{bmatrix}$ | 64×4×60×80 | 128×2×30×40 |
| Layer3 | $\begin{bmatrix} 3\times3\times3, 256, \text{stride } 2 \\ 3\times3\times3, 128 \end{bmatrix}$ | 128×2×30×40 | 256×1×15×20 |
| Avgpool | AvgPool3d | 256×1×15×20 | 256×1×1×1 |
| Reshape | View | 256×1×1×1 | 256 |
| Fc | 9d-fc | 256 | 9 |

**Funding.** This work was supported by the NSFC under Grants U19A2054.

**Disclosures.** The authors declare no conflicts of interest.

**Data availability.** Data underlying the results presented in this paper are not publicly available at this time but may be obtained from the authors upon reasonable request.